# fabSAM: A Farmland Boundary Delineation Method Based on the Segment Anything Model

Yufeng Xie[a,1], Hanzhi Wu[b,1], Hongxiang Tong[c], Lei Xiao[d], Wenwen Zhou[e], Ling Li[b,*], Thomas Cherico Wanger[f,g,*]

[a] *Department of Mathematics, Hong Kong University, Hong Kong, China*
[b] *Key Laboratory of Coastal Environment and Resources of Zhejiang Province, School of Engineering, Westlake University, Hangzhou, Zhejiang, China*
[c] *Alibaba Cloud, Hangzhou, Zhejiang, China*
[d] *Alibaba Group, Hangzhou, Zhejiang, China*
[e] *Sungrow iCarbon, Hefei, Anhui, China*
[f] *Sustainable Agricultural Systems and Engineering Laboratory, School of Engineering, Westlake University, Hangzhou, Zhejiang, China*
[g] *Production Technology and Cropping Systems Group, Department of Plant Production, AgroScope, Nyon, Switzerland*

\* Corresponding authors: liling@westlake.edu.cn (Ling Li), tomcwanger@gmail.com (Thomas Cherico Wanger)

[1] Contributed equally to this work.

**Abstract**

Delineating farmland boundaries is essential for agricultural management such as crop monitoring and agricultural census. Traditional methods using remote sensing imagery have been efficient but limited in generalization. The Segment Anything Model (SAM), known for its impressive zero-shot performance, has been adapted for remote sensing tasks through prompt learning and fine-tuning. Here, we propose a SAM-based **fa**rmland **b**oundary delineation framework (**fab**SAM) that combines a Deeplabv3+-based Prompter and SAM. Also, a fine-tuning strategy was introduced to enable SAM's decoder to improve the use of prompt information. Experimental results on the AI4Boundaries and AI4SmallFarms datasets have shown that fabSAM has a significant improvement in farmland region identification and boundary delineation. Compared to zero-shot SAM, fabSAM surpassed it by 23.47% and 15.10% in mIOU on the AI4Boundaries and AI4SmallFarms datasets, respectively. For Deeplabv3+, fabSAM outperformed it by 4.87% and 12.50% in mIOU, respectively. These results highlight the effectiveness of fabSAM, which also means that we can more easily obtain the global farmland region and boundary maps from open-source satellite image datasets like Sentinel-2.

*Keywords:* fabSAM, Segment Anything Model, Farmland boundary delineation, Semantic segmentation, Prompt engineering,

## 1. Introduction

One of the fundamental and challenging tasks of smart and precision agriculture is obtaining the spatial distribution of agricultural parcels and extracting their accurate boundaries, which also are indispensable prerequisites for downstream tasks in agricultural management, such as crop monitoring and agricultural census. Compared with conventional methods such as field surveys, methods that utilize Remote Sensing (RS) imagery can be much more efficient and cost-effective, as there are a large amount of publicly available RS datasets with various spatial and temporal resolutions, and the imagery interpretation technique is rapidly developing [1, 2].

Depending on the delineation methods used, existing studies extracting farmland boundaries from RS imagery can be roughly divided into two categories: traditional computer vision-based [3, 4] and deep learning-based methods [5, 6]. The traditional ones can be further sub-categorized into edge-based [7] and region-based techniques [4, 8]. As deep learning-based methods have shown greater accuracy and stability than traditional ones, models based on Deep Convolutional Networks (DCN) such as FCN [9], UNet [10], Deeplab [11] and their variants [5, 12] and Vision Transformers (ViTs) [13] have become popular recently.

However, the lack of training data in some countries and regions limits the generalization of these data-driven models [13]. To overcome this limitation, Visual Fundamental Models (VFMs) have been extended to RS imagery interpretation [14, 15, 16, 6], especially the Segment Anything Model (SAM) [17]. SAM has been trained and used on a dataset with over one billion masks and eleven million images for segmentation and therefore performs well on different kinds of tasks, including edge detection [17]. Also, SAM can be prompted by multiple forms of prompts, including points, bounding boxes, texts and masks, then allows it to be conveniently applied to specific downstream tasks [18].

Recently, a hybrid architecture has gained prominence in the RS community [14]. It contains a DCN model designed for object detection tasks to output bounding boxes that pinpoint the region of interest (ROI) in the RS images. Then, these boxes can serve as prompts for the SAM's prompt encoder. Combining the Prompter, a SAM-based workflow that automatically processes RS images has been developed.

Considering the variety of spatial resolution of RS datasets and the expected outputs of farmland boundary delineation tasks, we identified two challenges in constructing this hybrid model based on SAM: (1) How to generate the most effective prompts for SAM? (2) How can the SAM be fine-tuned to get more accurate outputs?

To address the aforementioned challenges, we released a SAM-based **fa**rmland **b**oundary delineation hybrid framework (**fab**SAM). This framework contains a Deeplabv3+-based Prompter that generates low-resolution masks and points as prompts, and two SAM-based parts that can separately identify the regions and boundaries of farmland. We then introduced a fine-tuning strategy for different delineation tasks. At the end, we evaluated whether fabSAM can improve the performance of the original Prompter on the AI4Boundaries (AI4B) and AI4SmallFarms (AI4S) datasets [19, 20]. To the best of our knowledge, this work is the first to introduce a hybrid architecture including a mask-prompt generator and a SAM-based block for farmland boundary delineation.

The rest of this article is organized as follows: Section **2** offers a review of SAM's applications in RS tasks, especially the farmland boundary delineation. Section **3** introduces the proposed fabSAM in detail. Section **4** presents the experimental results of fabSAM on the AI4B and AI4S datasets. Finally, the conclusion is given in section **5**.

## 2. Related works

Currently, there are mainly three kinds of methods that have been adopted to enhance SAM's performance on RS tasks: direct use of SAM, prompt learning skills, and fine-tuning techniques. In a case study from Bihar, India, SAM was employed directly for farmland boundary delineation in an unsupervised manner and successfully identified approximately 58% of the boundaries [13]. Also, SAM was directly applied to improve the Cropland Data Layer of the United States Department of Agriculture by incorporating a pixel-based classifier [21]. Besides, serving as a plug-in-and-play module, SAM was included in a CocoaNet architecture (i.e. a consistency-constrained multi-class attention model), which was used to obtain image-level class labels for RS images [22].

Prompt learning skills usually determine the performance of SAM through the quality of prompts, and how to automatically generate suitable taskoriented prompts has been systematically studied[16, 23]. A Python package for segmenting geospatial data with SAM was released, which allows users to prompt SAM with points, boxes and texts [24, 6]. Point and box prompts are the most widely adopted due to their flexibility, clarity and ease of acquisition. In a case study from Heilongjiang, China, a point prompter that focused on the ROI extracted by a simple machine learning-based classifier was added to SAM for crop field boundary delineation [18]. GeoSAM also proposed a CNN-based sparse prompter to generate point prompts for SAM to automatically segment mobility infrastructure [25]. Besides point prompts, a SAM-based model prompted by boxes that were generated from pseudo-labels was integrated into a CS-WSCDNet for the change detection task [26]. In addition, to take both point and box prompts into consideration, a SAM-based model for the local RS segmentation task was designed to be prompted by either center points of the ROI or bounding boxes that contain the ground-truth masks [15, 27]. As for text prompts, Text2Seg developed three basic methods to guide SAM for RS imagery segmentation: "Grounding DINO + SAM", "CLIP Surgery + SAM" and "SAM + CLIP" [28].

There are few studies focusing on mask prompts, however, they still have great potential in efficiently guiding SAM for RS segmentation compared to sparse prompts (i.e., points and boxes). First, mask prompts can provide the most exact spatial information for objects of interest. For example, GeoSAM contained a CNN-based mask prompter that can provide dense prompts for SAM[25]. Second, SAM supports self-guided iterative predictions. To be more specific, SAM can take the masks it predicted as input and iteratively predict masks to extract maximal information, then get stable and exact predictions. Take the Few-shot Self-guided Large Vision Model (Few-shot SLVM) as an example, this framework introduced an automatic prompt learning technique using the SAM for rendering coarse pixel-wise prompts instead of heavy reliance on manual guidance [29].

Recently, fine-tuning techniques have been proven to be necessary for extending SAM to downstream tasks [6, 30]. For example, several works have utilized a low-rank adaptation (LoRA) approach for crop-specific feature extraction and image encoder's fine-tuning [30, 31]. Also, there are some text-based one-shot learning approaches have been developed to retrain SAM [28]. Furthermore, a fine-tuning strategy of SAM's mask decoder was introduced to improve the performance of GeoSAM [25]. Despite all the efforts of integration and improvements of SAM in farmland boundary delineation, an integrated and fine-tuned approach of a mask Prompter and SAM-based block is underutilized.

## 3. Methodology

The overall framework of fabSAM is illustrated in Figure. 1. fabSAM involves two main blocks, including a Prompter and a SAM-based block. The RS images are first utilized by the Prompter to generate both mask and point prompts for SAM to localize farmland region. Following this, in the SAMbased block, the frozen SAM-based image encoder is utilized to extract highquality semantic features from original RS images, and the prompt encoder is employed to obtain explicit positional knowledge from prompts generated by the Prompter. Then, two fine-tunable decoders are introduced for farmland identification and boundary delineation tasks. Finally, the region mask and boundary mask are generated by the corresponding decoder and postprocessed to generate farmland parcels with exact and closed boundaries. This section first gives an introduction to the Prompter and the SAM-based block, followed by a discussion of our training strategies.

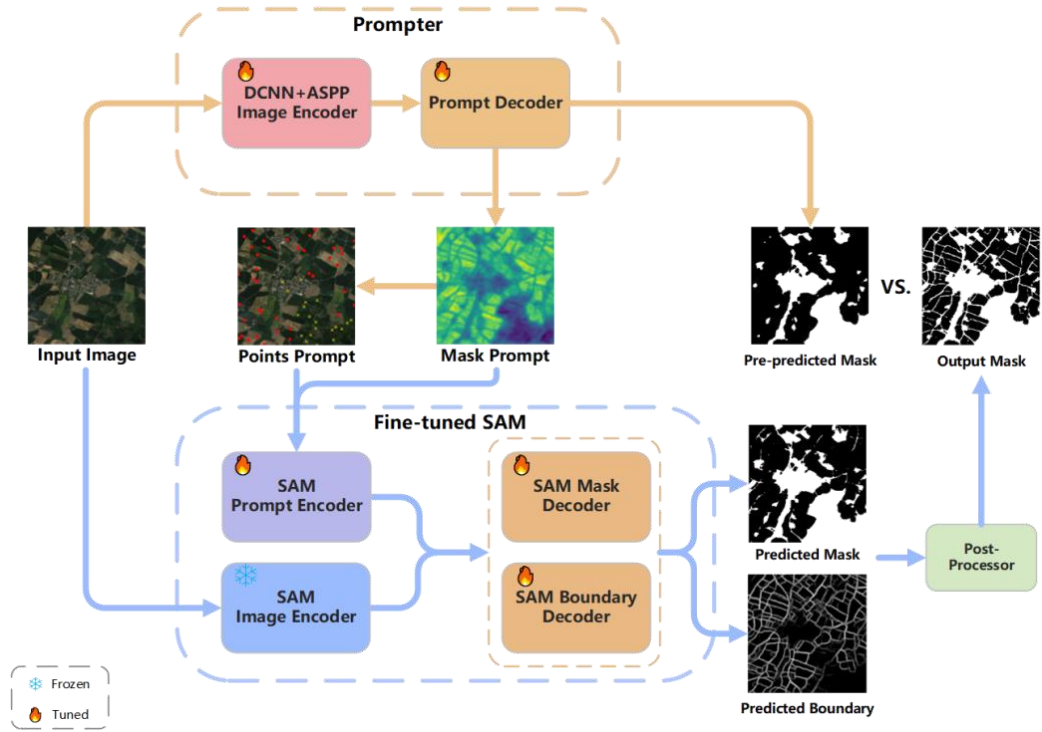

Figure 1: The overview of fabSAM framework consists of a Prompter and a SAM-based block (core block). The Prompter based on deeplabv3+ focuses on generating mask and point prompts for SAM (large yellow dashed box and arrows). In the SAM-based block (blue dashed

box and arrows), the input image and prompts are processed by the image encoder and prompt encoder. Corresponding embeddings are then utilized by two fine-tunable decoders to produce the masks for farmland identification and boundary de-lineation, respectively.

*3.1. Prompter*

The Prompter is designed to determine whether the pixels in the RS images belong to the farmland or background and generate both masks and points as prompts for the SAM-based block. As the range of farmland area is large and the boundaries are irregular, we use Deeplabv3+ [32] here as Prompter to better capture the multi-scale semantic features for farmland identification.

The Prompter follows the standard design of Deeplabv3+, which is an encoder-decoder structure. The encoder consists of two blocks: a ResNet50based DCNN block [33] and an Atrous Spatial Pyramid Pooling (ASPP)based block. In the prompt decoder, the low-level features drawn from the ResNet50-based DCNN block and the multi-level features extracted from the ASPP-based block are both aggregated and fed into a 3 × 3 convolution layer. In the end, an upsampling layer is used to restore the input shape. It is worth mentioning that the depthwise separable convolutions are used in the ASPP-based block and the prompt decoder to reduce the computation complexity [32].

Following this, we directly take the logits containing abundant location information output by the prompt decoder as the mask prompt, which follows the original design of SAM. Let $mp \in \mathbf{R}^{H \times W}$ denote the mask prompts, where $H$ and $W$ represent the resolution of the input RS image. Then, we introduce a point prompt generator (*Gen*) to obtain point prompts. In *Gen*, a farmland probability map ($P \in \mathbf{R}^{H \times W}$) is first produced by applying the sigmoid function on *mp*. Several foreground and background points are randomly chosen as prompts according to the following rules: (1) probability of belonging to farmland $P_i$ of the foreground pixel *i* must be larger than 0.7 while $P_j$ of background pixel *j* must be less than 0.3 to ensure the accuracy (these two thresholds are selected based on the experiment results); (2) the larger the $P_i$ (the lower the $P_j$), the more likely it is to be selected as point prompts. The whole process of the point prompt generation can be defined as:

$$P = Sigmoid(mp),$$
$$pp = Gen(P) \tag{1}$$

where $pp \in \mathbf{R}^{N\times 2}$ denote the point prompts, and *N* denotes the number of chosen points.

*3.2. SAM-based block*

The SAM-based block comprises four components that are duplicated from SAM: an image encoder ($Enc_I$), a prompt encoder ($Enc_P$) and two identical mask decoders ($Dec_M$ and $Dec_B$) that are fine-tuned separately for region identification and boundary delineation. The $Enc_I$ based on MAE pre-trained ViT [17] is used to extract high-level semantic features from the input images. Meanwhile, $Enc_P$ is utilized to encode the masks and points that are automatically generated by the Prompter as prompt embeddings. Then, the two light-weight decoders $Dec_B$ and $Dec_M$ based on self-attention and cross-attention are utilized to interact between image embedding and prompt embeddings and predict the boundary and region masks. To provide more insight, the whole process of the SAM-based block is defined as:

$$F_I = Enc_I(I),$$

$$F_{mp} = Enc_P(mp),$$

$$F_{pp} = Enc_P(pp), \quad (2)$$

$$B = Dec_B(F_I + F_{mp}, F_{pp}),$$

$$M = Dec_M(F_I + F_{mp}, F_{pp})$$

Where $F_I \in \mathbf{R}^{h \times w \times c}$, $F_{mp} \in \mathbf{R}^{h \times w \times c}$, and $F_{pp} \in \mathbf{R}^{N \times c}$ represent the image embeddings, mask representation, and point prompt tokens respectively, $h$ and $w$ represent the resolution of the image features, and $c$ denotes the feature dimension. Furthermore, $B$ and $M$ denote the predicted boundaries and regions.

A post-processor is then introduced to generate the farmland parcels with closed and exact boundaries. First, a symmetrical difference is made between the two outputs, $B$ and $M$, to more clearly distinguish the boundaries, and then the images are stitched into the original shape to get the final outputs.

### 3.3. Training strategies

Note that most of the classical segment models can be re-molded as autoprompt generators, and we can, therefore, treat the SAM-based model as a post-processing block that can improve the performance of these segment models. To make our framework more convenient to use and transform, we fine-tune the Prompter and the SAM-based block, separately.

#### 3.3.1. Training Prompter

In this training phase, the deeplabv3+-based Prompter calculates the loss by comparing the output pre-predicted mask to the extent mask of farmland. It is important to highlight that we add an auxiliary head after the image encoder to calculate the auxiliary loss, which is used as an assist in the training of a ResNet50-based DCNN block to optimize shallow layers [34, 35]. To be more detailed, the auxiliary head consists of a convolution layer with kernel size 3, a batch normalization layer and a convolution layer with kernel size 1 in series.

For the loss function to train the Prompter, $L_P$, we choose Cross-Entropy Loss ($L_C$) for both main loss ($L_m$) and auxiliary loss ($L_a$), so that they can be expressed using the same formula:

$$L_C = \frac{1}{N} \left( -\sum_{i=1}^{N} (\overline{y}_i^s \log y_i + (1 - \overline{y}_i^s) \log(1 - y_i)) \right)$$
$$L_P = w_m L_m + w_a L_a$$
(3)

Where $N$ denotes the number of pixels in the input image, $y^s_i$ denotes the ground truth of farmland region, $w_m$ and $w_a$ are the weight of main loss and auxiliary loss, and $y_i$ represents the predicted probability of pixel $i$. $y_i$ is generated by a Softmax function operating on logit, which is output by either a prompt decoder or an auxiliary decoder.

*3.3.2. Fine-tuning SAM-based block*

During the fine-tuning phase, we mainly update the parameters of the lightweight mask decoder and boundary decoder separately while the heavyweight image encoder is kept frozen. Meanwhile, we also fine-tune the parameters of the prompt encoder to better map the mask and point prompts to embeddings. For the loss function used to fine-tune the SAM-based block, we adopt the combination of Dice Loss and Focal Loss [25], which are both designed for situations where there is a strong imbalance between positive and negative samples. Dice Loss based on the dice coefficient is closely associated with the F1 score which is a widely used evaluation metric. The Dice Loss ($L_D$) can be articulated as:

$$L_D = 1 - \frac{2\sum_{i=1}^{N} \bar{y}_i^b y_i^O}{\sum_{i=1}^{N} \bar{y}_i^b + \sum_{i=1}^{N} y_i^O} \quad (4)$$

Where $y^b_i$ denotes the ground truth of farmland boundary or region, $y_i^O$ represents the output of the mask decoder, which is processed by a Sigmoid function.

The Focal Loss is an enhancement of the Cross-Entropy Loss in terms of the imbalance of samples. Two parameters $\alpha$ and $\gamma$ are introduced to make the model focus on fewer examples and difficult-to-distinguish examples. The definition of Focal Loss ($L_F$) is shown as:

$$L_F = \begin{cases} -\alpha(1-y_i^O)^\gamma \log y_i^O, & \bar{y}_i^b = 1 \\ -(1-\alpha)(y_i^O)^\gamma \log(1-y_i^O), & \bar{y}_i^b = 0 \end{cases} \quad (5)$$

In conclusion, the loss function for fine-tuning the SAM-based block ($L_{FT}$) can be expressed as:

$$L_{FT} = w_D L_D + w_F L_F \quad (6)$$

Where $w_D$ and $w_F$ denote the weight of Dice Loss ($L_D$) and Focal Loss ($L_F$) respectively.

## 4. Experiments and discussion

### 4.1. Datasets and pre-processing

Two open AI-ready field boundary datasets named AI4B [19] and AI4S [20] were utilized to fine-tune and evaluate fabSAM. For the AI4B dataset, these images were captured in 2019 over Austria, Catalonia, France, Luxembourg, the Netherlands, Slovenia and Sweden. We employed the 10-meter cloud-free Sentinel-2 subset, which consisted of 7831 monthly composite images captured in May, and their corresponding ground truth boundary labels contained an extent mask and a boundary mask. For the AI4S dataset, 62 cloud-free Sentinel-2 images with a spatial resolution of 10 meters were provided along with extent labels in vector format and boundary labels in raster format. The areas in these images were distributed over Vietnam and Cambodia.

Before taking these images as input for fabSAM, we pre-processed them as follows. First, we converted the data from both two datasets to Geotiff format, and performed band rendering and contrast enhancement - in which minimum and maximum band values were set to 0 and 3000 respectively - to make sure the enhanced images can be shown clearly. Then, we exported the rendered images (band range 0-255) and cropped them to a size of 256 × 256 pixels. As for the label data, the extent masks of farmland (transformed into raster format) were used for the training of Prompter and the fine-tuning of the mask decoder. The boundary masks were used for the fine-tuning of the boundary decoder. For both datasets, we took 70% of the data for training, 15% for testing, and the remaining 15% for validation.

*4.2. Implementation details*

Our proposed model was developed using the Pytorch framework. MM-Segmentation and Lightning packages were used to train and fine-tune the Prompter and the SAM-based block separately. All model training and experiments were conducted using an NVIDIA A10 GPU with 24GB RAM.

Detailed parameters setting for the training of Prompter and SAM-based block are listed in Table. 1.

Table 1: Parameters setting for model training and fine-tuning

| Configuration | Prompter | SAM-based block |
| --- | --- | --- |
| Optimizer | SGD | Adam |
| Batch size | 8 | 4 |
| Total epochs (iterations) | 80000 iterations | 20 epochs |
| Initial learning rate | 0.004 | 0.0003 |
| LR policy | Poly | Poly |
| Decay factor | 0.0001 | 0.0001 |

*4.3. Evaluation metrics*

To clearly and quantitatively evaluate the performance of fabSAM on farmland identification and boundary delineation, we used four evaluation indicators: Intersection over Union (IoU), F1-score, Accuracy, and a composite indicator mean Intersection Over Union (mIOU). The calculation formulas for the first three metrics are as follows:

$$\text{IoU} = \frac{TP}{TP+FP+FN}$$
$$F1 = \frac{2TP}{2TP+FP+FN} \quad (7)$$
$$\text{Accuracy} = \frac{TP+TN}{TP+FP+TN+FN}$$

Where *TP*, *FP*, *TN*, and *FN* represent the numbers of pixels that belong to true positive, false positive, true negative, and false negative, respectively.

In order to measure the comprehensive ability of our fabSAM in the tasks of both region recognition and boundary delineation, we introduced a composite indicator mIOU, which is defined as follows:

$$\text{mIOU} = \frac{1}{2}(IoU_r + IoU_b) \qquad (8)$$

Where $IoU_r$ and $IoU_b$ denote the $IoU$ of the farmland region class and boundary class, respectively. The extent of farmland fragmentation and the resolution of RS images can influence the mIOU value. In our experiment, a model with a mIOU value exceeding 35% was deemed to exhibit satisfactory performance, while a value surpassing 50% indicated superior performance.

*4.4. Results and discussion*

*4.4.1. Performance comparison*

In order to demonstrate the performance enhancement achieved by fabSAM compared to the original prompter and SAM, we initially conducted a comparative analysis of fabSAM against zero-shot SAM [17] and Deeplabv3+ [32]. Additionally, we selected two state-of-the-art semantic segmentation models - UNet+PSPNet (based on convolution) [36]) and MaskFormer (based on transformer architecture) [37]) - for further comparison to verify the effectiveness of fabSAM. We then quantitatively compared the performance of fabSAM with these models based on IoU, F1-score, Accuracy and mIOU evaluation metrics.

From Table. 2, our fabSAM exhibited significant advancements over other models in the composite indicator mIOU on both AI4B and AI4S datasets. For example, compared to zero-shot SAM, fabSAM surpassed it by 23.47% and 15.10% in mIOU on AI4B and AI4S datasets, respectively. Also, fabSAM surpassed Deeplabv3+ by 4.87% and 12.50% in mIOU, respectively.

While showing only slight improvements in some metrics, for example, fabSAM outperformed MaskFormer by only 0.31% in Accuracy on farmland region identification on the AI4B dataset. However, fabSAM truly shined in the boundary delineation task, outperforming all other models on the AI4B dataset in IoU, F1 and Accuracy. Overall, fabSAM has superior comprehensive performance, particularly excelling in farmland boundary delineation.

Table 2: Quantitative comparison on the performance of fabSAM and selected models: Zero-shot SAM, Deeplabv3+, UNet+PSPNet, MaskFormer

| Dataset | Method | Farmland Region | | | Farmland Boundary | | | Composite |
|---|---|---|---|---|---|---|---|---|
| | | IoU | F1-score | Accuracy | IoU | F1-score | Accuracy | mIOU |
| AI4B | **fabSAM** | **60.64** | **71.11** | **85.66** | **37.40** | **51.01** | **84.63** | **49.02** |
| | SAM (zero-shot) | 44.23 | 57.22 | 74.69 | 6.86 | 12.37 | 75.16 | 25.55 |
| | Deeplabv3+ | 59.30 | 69.85 | 85.86 | 28.99 | 40.87 | 83.66 | 44.15 |
| | Unet+PSPNet | 57.98 | 68.54 | 84.80 | 29.39 | 41.41 | 84.15 | 43.69 |
| | Maskformer | 57.93 | 68.75 | 85.35 | 31.97 | 44.73 | 82.94 | 44.95 |
| AI4S | **fabSAM** | **84.93** | **91.52** | **88.47** | **27.62** | **42.84** | **72.51** | **56.28** |
| | SAM (zero-shot) | 73.95 | 84.70 | 79.76 | 8.40 | 15.36 | 69.84 | 41.18 |
| | Deeplabv3+ | 84.30 | 91.12 | 87.92 | 3.25 | 5.89 | 74.28 | 43.78 |
| | Unet+PSPNet | 83.07 | 90.34 | 86.94 | 8.13 | 13.61 | 70.76 | 45.60 |
| | Maskformer | 83.30 | 90.48 | 87.17 | 27.23 | 42.36 | 31.36 | 55.27 |

The visual performance of fabSAM on several test images is shown in Figure. 2. These figures demonstrate that the performance of fabSAM in recognizing small-scale farmland regions and irregular boundaries was better than zero-shot SAM, Deeplabv3+ and other models. In order to see more intuitively how fabSAM can optimize the outputs of the Deeplabv3+ model, which also serves as the primary component of the Prompter in fabSAM, the logit maps generated by deeplabv3+, fabSAM's mask decoder and boundary decoder are presented in Figure. 3. These maps represent the probability that corresponding pixels belong to the foreground. Focusing on the areas delineated by the red boxes in Figure. 3, we can find that fabSAM outperforms Deeplabv3+ in terms of differentiating farmland regions from complex backgrounds, including water bodies, roads, and trees. Based on the multilevel features extracted by SAM-based Image Encoder, fabSAM has much better prediction stability by increasing the difference between foreground and background in logit maps.

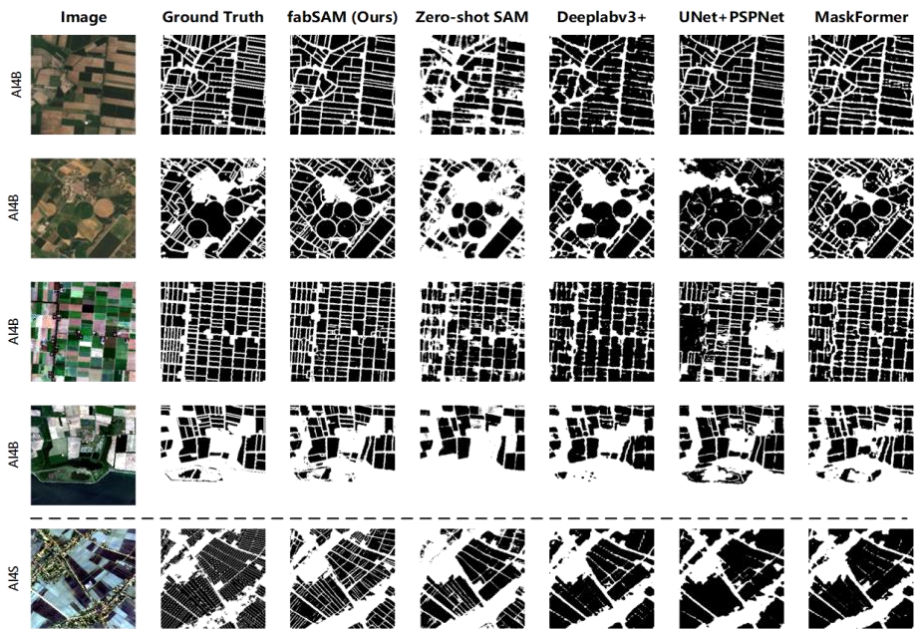

Figure 2: Qualitative comparison on farmland boundary delineation: fabSAM, Zero-shot SAM, Deeplabv3+, UNet+PSPNet and MaskFormer.

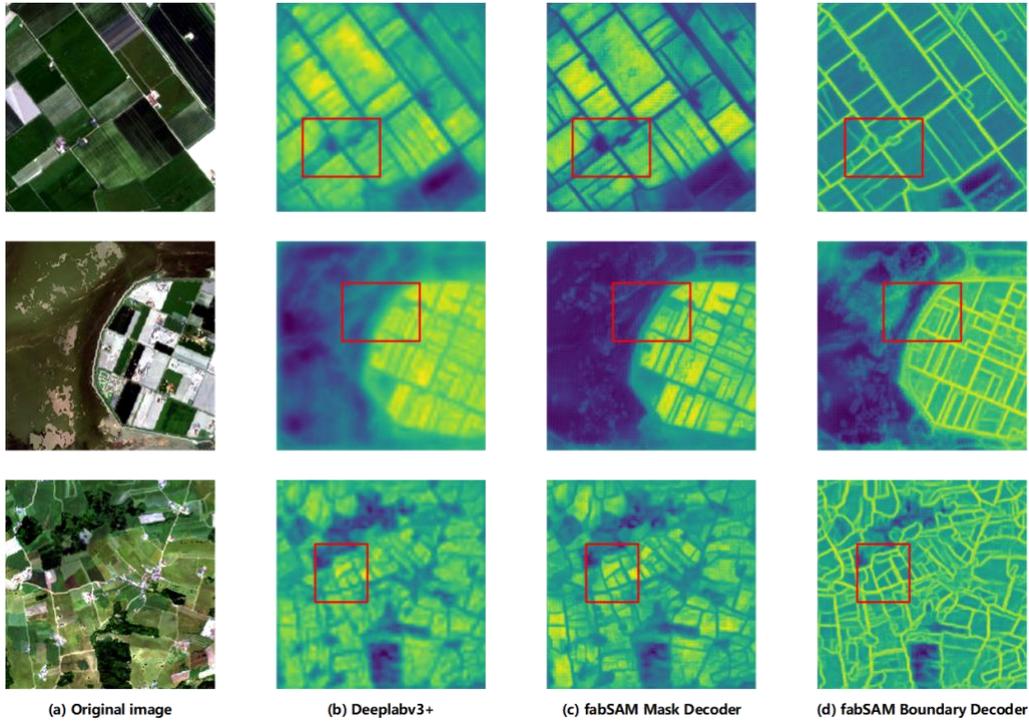

Figure 3: Improvement in prediction accuracy of logits: (a) original image, (b) logits predicted by Deeplabv3+, (c) logits predicted by fabSAM's Mask Decoder, (d) logits predicted by fabSAM's Boundary Decoder.

As for fabSAM's performance on the farmland boundary delineation task, Figure. 4 shows the comparisons between fabSAM and the best-performing three models on the AI4S dataset: Deeplabv3+, MaskFormer and UNet+PSPNet. The boundaries generated by fabSAM are more accurate and have better connectivity. Though the clarity of these images in the dataset was affected by the resolution and acquisition time of the satellite imagery, fabSAM still performs well in difficult-to-distinguish regions.

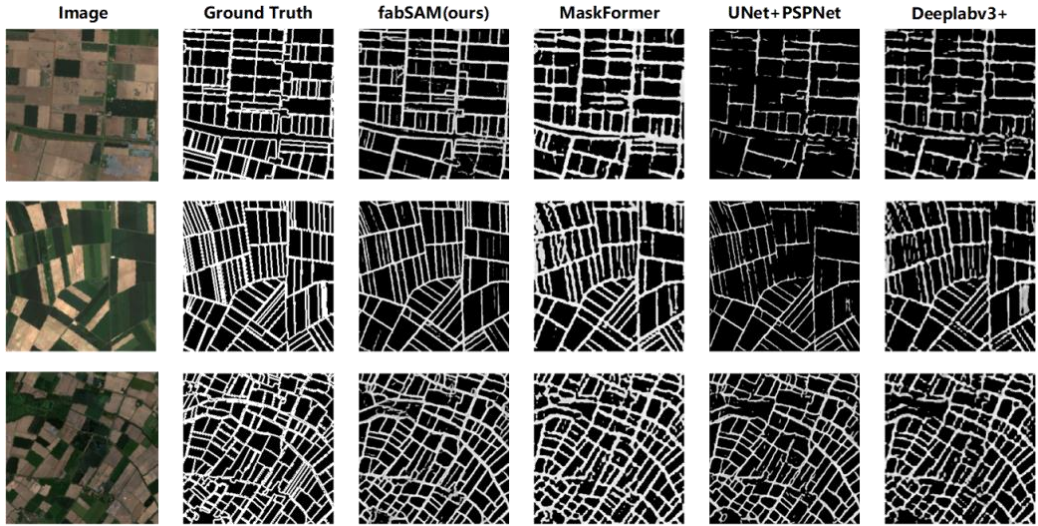

Figure 4: Qualitative comparison on boundary delineation: fabSAM, MaskFormer, UNet+PSPNet and Deeplabv3+.

*4.4.2. Ablation experiments*

In this part, we removed a certain part of fabSAM to verify the effectiveness of each component, and the quantitative results of different settings are shown in Table. 3. The first row corresponds to the original fabSAM, which has the highest scores. Notably, not fine-tuning the decoder would lead to a substantial decline in model performance, with a 16.32% IoU decrease for region identification and a 38.65% F1 decrease for boundary delineation. The effects of different forms of prompts are shown in the last two rows. Considering the both region identification and boundary delineation tasks, removing the Mask Prompt resulted in an approximately 1% decrease in IoU and F1 while removing the Point Prompt caused an approximately 2% decrease.

Table 3: Ablation experimental results. FTD = Fine-tuning Decoder, FTPE = Fine-tuning Prompt Encoder, MP = Mask Prompt, PP = Point Prompt

| Techniques | | | | | Region | | Boundary | |
|---|---|---|---|---|---|---|---|---|
| SAM ENC | FTD | FTPE | MP | PP | IOU | F1 | IOU | F1 |
| ✓ | ✓ | ✓ | ✓ | ✓ | **60.55** | **71.04** | **37.4** | **51.01** |
| ✓ | × | ✓ | ✓ | ✓ | 44.23 | 57.22 | 6.85 | 12.36 |
| ✓ | ✓ | × | ✓ | ✓ | 60.44 | 70.82 | 36.85 | 50.52 |
| ✓ | ✓ | ✓ | × | ✓ | 59.58 | 69.96 | 37.26 | 50.57 |
| ✓ | ✓ | ✓ | ✓ | × | 58.11 | 69.04 | 35.85 | 49.11 |

Although fabSAM performs well on the AI4B and AI4S datasets, this framework has a notable limitation that its accuracy depends mainly on the accuracy of the Prompter. In addition, as SAM itself has the disadvantage of missing fine structures, it may lead fabSAM to ignore small areas of fragmented farmland. This is the reason why the improvement in the performance of fabSAM on region identification is not significant.

## 5. Conclusion

In this paper, we proposed a fabSAM framework for farmland region identification and boundary delineation tasks, which can generate farmland region parcels with valid perimeters and boundary maps with good connectivity from satellite images without human intervention. We also introduced a new architecture to generate prompts for the SAM from masks and points that an arbitrary semantic segmentation model predicts. Furthermore, a SAM-based block decoder was developed for two tasks: farmland region identification (instance segmentation) and boundary delineation (edge detection). Experimental results on the AI4B and AI4S datasets highlight the effectiveness and great generalization of fabSAM, which also means that we can more easily obtain the global farmland region and boundary maps from open-source satellite image datasets like Sentinel-2. However, according to the results of ablation experiments, it is challenging to improve segmentation performance by adding a SAM-based block and fine-tuning a lightweight decoder. Therefore, to promote the future development of smart and precision farmland planning and management, future work could focus on how to generate more stable and exact prompts, and how to construct new modules that refine boundary details.